\pdfoutput=1
\documentclass[lettersize,journal]{IEEEtran}
\usepackage{amsmath,amssymb,amsthm}
\usepackage{booktabs}
\usepackage{graphicx}
\usepackage{multirow}
\usepackage{xcolor}
\usepackage[colorlinks,linkcolor=blue,citecolor=blue,urlcolor=blue]{hyperref}
\usepackage{cite}
\usepackage[ruled,linesnumbered]{algorithm2e}
\usepackage{tikz}
\usetikzlibrary{positioning,fit,backgrounds,arrows.meta}

\newcommand{\R}{\mathbb{R}}

\newcommand{\cG}{\mathcal{G}}
\newcommand{\cS}{\mathcal{S}}
\newcommand{\cV}{\mathcal{V}}
\newcommand{\cE}{\mathcal{E}}
\newcommand{\cL}{\mathcal{L}}

\newcommand{\cQ}{\mathcal{Q}}

\newcommand{\cH}{\mathcal{H}}

\newtheorem{definition}{Definition}

\newcommand{\method}{LoGIC}
\newcommand{\gtfm}{G2T-FM}
\newcommand{\graphpfn}{GraphPFN}
\newcommand{\tabpfn}{TabPFNv2}

\begin{document}

\title{LoGIC: Budgeted Context Construction for\\
Node-Level Graph In-Context Learning\\
with Tabular Foundation Models}

\author{Mingqi~Yang, Zidong~Guo, Jihui~Yang, and~Wenming~Zuo
\thanks{This work has been submitted to the IEEE for possible
publication. Copyright may be transferred without notice, after which
this version may no longer be accessible.}%
\thanks{\emph{(Corresponding author: Wenming Zuo.)}}%
\thanks{Mingqi Yang, Zidong Guo, and Wenming Zuo are with the Department of
Electronic Business, South China University of Technology, Guangzhou,
Guangdong 510006, China (e-mail: yangmq@scut.edu.cn;
202520158576@mail.scut.edu.cn; wmzuo@scut.edu.cn).}%
\thanks{Jihui Yang is with the Meta Superintelligence Lab (e-mail:
jihui@meta.com).}%
\thanks{This work was supported in part by the National Natural Science
Foundation of China under Grant 62506134, in part by the
Guangdong--Hong Kong--Macao Applied Mathematics Center Project under Grant
2026A1515060008, and in part by the Guangdong Natural Science Foundation
under Grant 2026A1515010171.}}

\maketitle

\begin{abstract}
Tabular foundation models have become powerful graph learners. Systems such as \gtfm{} and \graphpfn{} encode each node as a feature row and make predictions through in-context learning (ICL), with labeled rows serving as the prompt. Current protocols employ the complete training table as context, causing attention to scale quadratically with the labeled pool and introducing preprocessing and memory bottlenecks. We investigate context construction for node-level graph ICL: which labeled nodes and auxiliary unlabeled nodes should constitute the prompt for specified queries. We formulate this allocation in terms of two resources: a labeled-context budget for predictive evidence and an unlabeled-halo budget for adapter message passing without using label capacity. We present \method{}, which retrieves labeled nodes via structural, feature-based, and coverage channels, shares each context across the queries in a graph-local cluster, incorporates an unlabeled halo for adapter backbones, and chooses the channel and context budget without test labels. Across three backbone configurations drawn from two model families on GraphLand, budgeted contexts maintain locally runnable full-context performance, stay competitive with published large-dataset results, and markedly lower peak memory requirements compared with full-context and whole-graph inference. They further permit frozen graph ICL on million-node graphs without retraining. Our analysis identifies when retrieval channels work best and connects their behavior with graph properties.
\end{abstract}

\begin{IEEEkeywords}
Graph foundation models, in-context learning, tabular foundation models, context retrieval, graph machine learning, scalability
\end{IEEEkeywords}

\section{Introduction}
\label{sec:intro}

\IEEEPARstart{F}{oundation} models pretrained on extensive data have transformed natural language processing and computer vision, and graph learning has sought a comparable modeling paradigm. Graphs render this transfer challenging: datasets across different domains seldom share feature spaces, degree distributions, or label semantics, and cross-graph transfer continues to be difficult~\cite{gft,graphany}. Tabular foundation models, pretrained for in-context learning (ICL) on synthetic tables, have become effective graph learners when every node is encoded as a feature row. \gtfm{}~\cite{g2tfm} builds these rows from raw attributes, neighborhood feature aggregates, and structural encodings, and subsequently employs a frozen prior-data fitted network (PFN)~\cite{tabpfn,tabpfnv2} for node prediction. \graphpfn{}~\cite{graphpfn} alternatively enhances the tabular backbone with pretrained graph-attention adapters. Across varied industrial benchmarks, PFN-based graph models have been shown to equal or surpass well-tuned GNNs on multiple datasets~\cite{faireval2026}.

The prompt employed by these systems has received less consideration than the row representation or the backbone. In ICL, the labeled rows within the prompt provide the model's supervision during inference, but both \gtfm{} and \graphpfn{} employ the entire training table as context. \graphpfn{} observes that its ``current implementation requires processing the entire dataset at once, which leads to significant memory consumption''~\cite{graphpfn}, and lists subgraph sampling as future work. G2T-TabPFNv2 encounters a similar limitation: practical memory consumption increases with both context size and feature width, and the published GraphLand protocol consequently uses PCA when full-feature preprocessing surpasses available memory~\cite{g2tfm}. TAG~\cite{tag} limits the prompt by randomly subsampling labeled rows, although random subsampling regulates capacity without determining which rows are pertinent to the current queries.

The full-table default is likewise costly. Attention across the context scales quadratically with the number of labeled rows, and an identical cost reappears for every prediction batch. For adapter-based backbones, peak memory additionally covers the resident graph tokens, meaning that it depends on the graph, its features, and the execution configuration rather than solely on node count. In the matched local measurements presented in Section~\ref{sec:rq2}, whole-graph \graphpfn{} consumes 19.59\,GB on the 50k-node artnet-exp graph and 7.71\,GB on the 57k-node city-roads-M graph, whereas the respective \method{} configurations consume 8.24\,GB and 5.73\,GB. The concern is thus not a constant node-count threshold. Rather, full-graph residency is linked to the input graph, while a retrieved-subgraph interface links residency to explicit query, labeled-context, and halo budgets. This difference becomes increasingly important as graph size expands.

This paper consequently asks: \textit{given a frozen tabular backbone and a budget, which labeled nodes, and which auxiliary unlabeled nodes, should constitute the prompt for a specified set of queries?}

Current selection techniques do not directly address this question. Retrieval-augmented tabular ICL ranks rows in feature space ~\cite{localpfn,tabdpt,mixturepfn} and disregards graph structure. Demonstration selection for graph ICL has been investigated primarily with LLM backbones, frequently employing only a small number of demonstrations chosen through global centrality, learned retrieval, or query-dependent neighborhoods ~\cite{graphicl,askgnn,graphprompter,grail}. In GraphICL's own ablation, PageRank-based selection fails to surpass random selection~\cite{graphicl}. Neither research direction encompasses PFN-based graph FMs, in which the context may include thousands of labeled rows and, for adapter backbones, has to be provided as a subgraph instead of an unordered collection of rows. Section~\ref{sec:related} examines these directions in detail.

We formalize context construction as budgeted allocation across two resources: $k$ labeled context rows and $h$ unlabeled \emph{halo} nodes. The labeled budget governs the supervised evidence, whereas the halo provides unlabeled nodes for adapter message passing without using label capacity. We subsequently introduce \method{} (locally clustered graph in-context construction), a retrieval approach with three scoring channels: personalized-PageRank locality, feature-space similarity, and coverage sampling. \method{} shares each retrieved context among all queries in a graph-local cluster, thereby amortizing retrieval and encoding over several queries. The retrieval channel and context budget are regarded as configuration choices and are chosen without test labels. This design complements advances in graph tabular foundation models: it neither replaces nor retrains the backbone, but offers a shared inference interface that bounds the labeled context and, where necessary, the supporting subgraph.

\begin{figure*}[t]
\centering
\includegraphics[width=\textwidth,trim=0 180bp 0 0,clip]{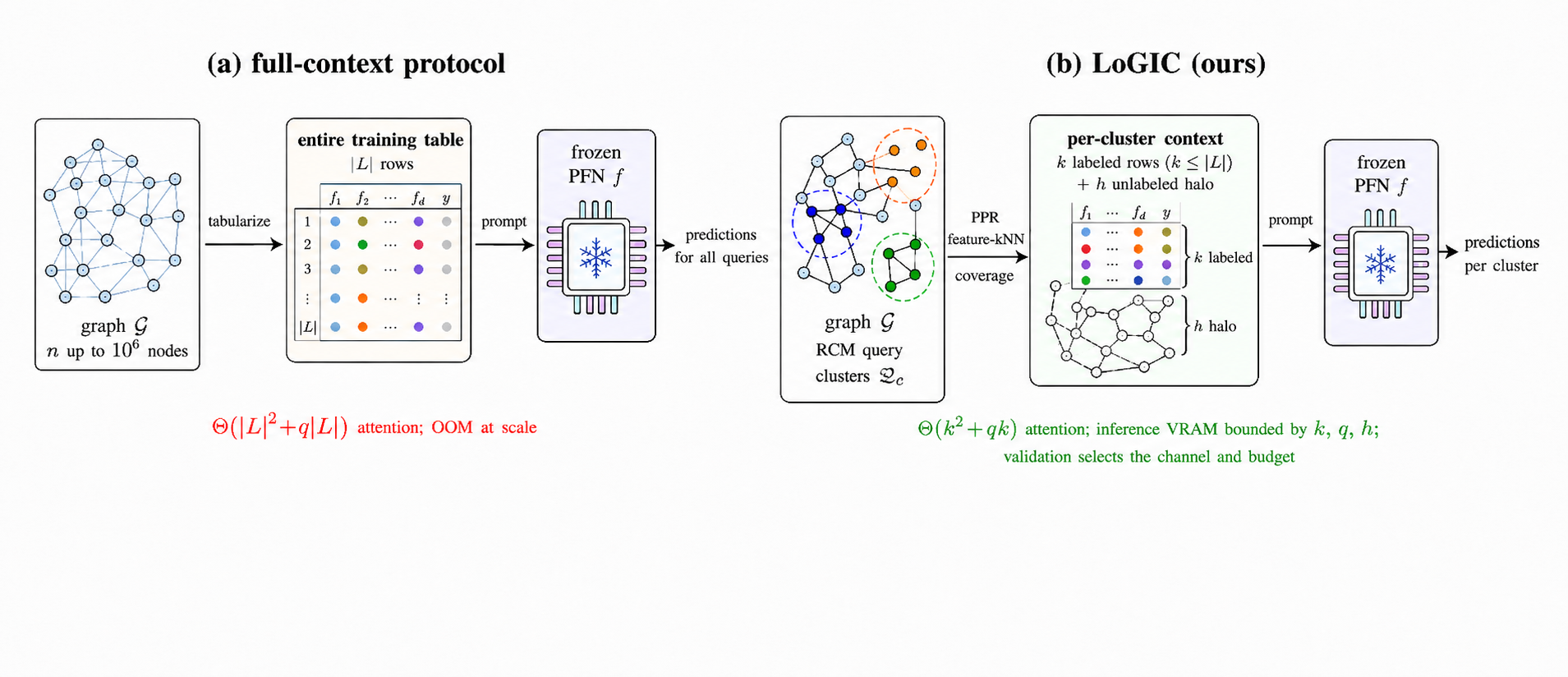}
\caption{Full-context graph ICL (a) versus budgeted context construction (b). \method{} retrieves a labeled context per graph-local query cluster through three channels and adds an unlabeled halo for adapter backbones.}
\label{fig:overview}
\end{figure*}

Our experiments on GraphLand~\cite{graphland} encompass three backbone configurations from the pure-tabular and graph-adapter families, three retrieval channels, and budgets spanning hundreds to tens of thousands of rows. The highest-performing channel differs across datasets. Graph-local retrieval tends to achieve the best performance when labels are smooth over edges, while coverage sampling is preferred when edge-label assortativity is distinctly negative. These findings suggest that the suitable retrieval channel depends on both the graph and the accessible context budget. Since peak accelerator memory during backbone inference is constrained mainly by the query, labeled-context, and halo budgets, the same construction supports configurations unavailable under the respective full-context or full-graph protocols: full-feature inference where the official protocol needs PCA truncation, \graphpfn{} inference that surpasses its published full-graph accuracy with bounded subgraph residency, and frozen graph ICL on a 1.6-million-node graph. Figure~\ref{fig:overview} compares the two protocols, and the experiments associate retrieval-channel behavior with the context budget and measurable graph properties.

The primary contributions of this paper are summarized as follows:
\begin{itemize}
\item We recognize and formulate context construction for node-level graph ICL as a budgeted allocation problem involving two resources, labeled context rows and unlabeled halo nodes.
\item We introduce \method{}, a context-construction framework developed to accommodate multiple graph tabular foundation model inference interfaces, incorporating multi-channel retrieval, cluster-shared contexts, and an unlabeled halo for graph-adapter backbones. It modifies neither backbone parameters nor training and is implemented for both row-based and graph-adapter inference.
\item We empirically connect retrieval-channel behavior to the context budget and measurable graph properties.
\item We perform an extensive evaluation across eight GraphLand datasets and three backbone configurations covering two families, examining predictive performance, efficiency, scalability, and controlled ablations under a consistent and reproducible protocol.
\end{itemize}

\section{Related Work}
\label{sec:related}

\subsection{Tabular Foundation Models and Context Retrieval}
TabPFN~\cite{tabpfn,tabpfnv2} positioned prior-data fitted networks as a training-free framework for tabular prediction, and subsequent models including TabICL~\cite{tabicl}, LimiX~\cite{limix}, and TabPFN-3~\cite{tabpfn3} have increased the supported context size from thousands to millions of rows. A complementary research direction examines what should be included in the context. LoCalPFN ~\cite{localpfn} obtains a kNN context for each query and demonstrates on synthetic tables that a local context can surpass the full table when the complete table fits the prompt. For real tables, however, row-cap restrictions preclude the equivalent full-context comparison. TabDPT~\cite{tabdpt} integrates retrieval into pretraining, MixturePFN~\cite{mixturepfn} directs queries to prompt experts, chunked attention~\cite{chunkedtabpfn} lowers the cost of long contexts, and CRUMB~\cite{crumb}, concurrent with this work, groups queries into clusters and shares a distribution-matched context within each cluster. LimiX further provides a retrieval-based ensemble, although it continues to pass over every sample ~\cite{limix}. These approaches motivate context selection for frozen tabular models, yet their selection signals are tabular: rows are evaluated in feature space or through distributional matching, rather than through graph structure. In the language-model setting, demonstration selection is similarly recognized as changing the behavior of frozen models; embedding-based retrieval~\cite{kate}, learned prompt retrievers~\cite{epr}, and an extensive follow-up literature ~\cite{iclsurvey} document substantial accuracy differences from selection alone.

Chunked attention and distribution-matched batching improve how a tabular PFN processes a provided context~\cite{chunkedtabpfn,crumb}. LoGIC addresses the prior decision: which labeled graph nodes should fill the bounded PFN context, and, for adapter backbones, which unlabeled nodes should maintain local receptive fields. The resulting context can be processed by a memory-efficient PFN implementation without modifying LoGIC's graph-aware selection interface.

\subsection{Tabular Foundation Models as Graph Foundation Models}
\gtfm{}~\cite{g2tfm} transforms node prediction into tabular ICL via neighborhood feature aggregation and structural encodings, while \graphpfn{}~\cite{graphpfn} extends a LimiX-based PFN backbone using pretrained adjacency-masked graph-attention adapters. Subsequent work employs tabular FMs for anomaly detection~\cite{tfm4gad}, link prediction with constructed contexts~\cite{tfmlinker}, and graph tabularization~\cite{tabpfngn}. NodePFN alternatively pretrains a graph-aware PFN with local message passing over synthetic graph priors~\cite{nodepfn}. An independent assessment describes PFN-based graph FMs as competitive with, and frequently stronger than, tuned GNNs across varied benchmarks~\cite{faireval2026}. These systems demonstrate that tabular FMs can function as graph foundation models, although their standard inference protocols continue to employ the complete training context; \graphpfn{} additionally identifies whole-graph processing as its primary limitation~\cite{graphpfn}. TAG~\cite{tag} constrains the prompt through class-balanced random subsampling of labeled rows, offering capacity control but no graph-aware retrieval policy. Earlier graph foundation models seek transfer through alternative routes: PRODIGY~\cite{prodigy} pretrains GNNs on prompt-graph tasks for few-shot in-context learning across graphs, whereas cross-domain models including GFT~\cite{gft} and GraphAny~\cite{graphany} transfer through architectures. Their context, if present, remains the entire graph.

\subsection{Selecting Nodes on Graphs}
Demonstration selection for graph ICL has been investigated with LLM backbones. GraphICL~\cite{graphicl} evaluates prompt designs for node classification and ranks demonstrations through random sampling, cosine similarity, or global PageRank centrality with no more than six demonstrations. It finds that the PageRank variant fails to exceed random selection. AskGNN~\cite{askgnn} trains a structure-enhanced retriever over GNN embeddings, GraphPrompter~\cite{graphprompter} jointly optimizes prompt generation, selection, and augmentation across stages, GRAIL~\cite{grail} obtains top-$k$ relevant nodes via GNN embeddings for LLM in-context learning, and RAGraph~\cite{ragraph} inserts retrieved subgraphs into a pretrained GNN using a prompt mechanism.

Selection subject to a node budget also occurs in graph active learning, which selects nodes for annotation using centrality and diversity~\cite{age}, feature propagation~\cite{featprop}, partitioning~\cite{graphpart}, or influence maximization~\cite{grain}. Regarding scalability, sampling for GNN training~\cite{graphsage,clustergcn,shadowgnn} regulates gradient variance across epochs, while graph condensation~\cite{gcond} generates a compact training set once for each graph. PPR itself has an extensive history as a propagation operator~\cite{appnp,pprgo}. Label propagation and the label-as-feature technique~\cite{cs,unimp} instead incorporate label information on the feature side.

Graph samplers including neighborhood sampling and Cluster-GCN build computational subgraphs for GNN training~\cite{graphsage,clustergcn}. Their sampled nodes chiefly regulate message-passing cost; they neither determine which labels should fill a limited PFN prompt nor distinguish labeled evidence from unlabeled structural support. LoGIC focuses on this frozen-inference interface. The mechanisms are nonetheless compatible: a graph partitioner could substitute for RCM in query clustering or supply candidate subgraphs, after which LoGIC would assign its label and halo budgets.

\subsection{Positioning}
Our setting diverges from these research directions in several ways. In contrast to tabular retrieval methods~\cite{localpfn,tabdpt}, our selection signals exploit the graph, and PPR is employed to retrieve from the labeled pool instead of propagating predictions~\cite{appnp,pprgo}. Relative to graph-ICL methods ~\cite{graphicl,askgnn,grail} and active-learning methods ~\cite{age,featprop,graphpart}, our backbone processes thousands of labeled rows simultaneously, the labels are already available, and adapter backbones demand that the retrieved context be delivered as a subgraph. The halo budget is established for this final requirement. Graph ICL with tabular FMs is further a setting where the full-context protocol is standard~\cite{g2tfm,graphpfn} and remains executable on smaller benchmarks, allowing the subset-versus-full comparison proposed by the tabular direction to be conducted directly; Section~\ref{sec:rq2} presents this comparison under matched backbones. To the best of our knowledge, budgeted two-resource context construction for frozen tabular graph FMs has not been examined previously. TAG's random subsampling~\cite{tag} constitutes the nearest prior mechanism, and we employ feature-kNN retrieval~\cite{localpfn,tabdpt} as the feature-space control in the channel ablations.

A full-context graph PFN~\cite{g2tfm,graphpfn} merges three decisions: which labeled rows support the current queries, how PFN attention arranges those rows, and which nodes and edges stay resident for adapter message passing. LoGIC handles the first decision and provides a halo budget for the third. Chunked TabPFN~\cite{chunkedtabpfn} alters the second, whereas CRUMB~\cite{crumb} combines query batching with distribution-matched tabular retrieval; neither exploits graph distance nor builds adapter subgraphs. By contrast, neighborhood sampling and Cluster-GCN ~\cite{graphsage,clustergcn} decrease graph computation but do not assign a labeled PFN prompt. These methods are not substitutes for one another, although they can be integrated: memory-efficient attention can process a LoGIC context, and graph partitioning can substitute for the RCM query order while preserving LoGIC's labeled-context and halo allocation.

This layered perspective also establishes the suitable comparison. Full-context and full-graph runs using the same backbone separate the effect of context construction, whereas fixed PPR, feature-kNN, and coverage policies separate the selection signal. An end-to-end comparison that simultaneously modifies the attention kernel, query batching rule, graph partition, and retrieved labels would confound these factors. An integrated system is nonetheless a promising direction: LoGIC can choose the labeled evidence and structural support, after which chunked attention or a distribution-matched scheduler can process the bounded context.

\section{Preliminaries and Problem Statement}
\label{sec:prelim}

\subsection{Graph In-Context Learning}
Let $\cG=(\cV,\cE,X)$ denote a graph comprising node features $X\in\R^{n\times d}$, a labeled subset $L\subset\cV$ with targets $y_L$, and a query set $\cQ\subseteq\cV\setminus L$ (transductive node classification or regression). A \emph{tabularized} graph FM transforms every node into a row representation $r(v)$ and subsequently applies a frozen PFN backbone $f$ through in-context learning. In \gtfm{}, the row includes raw attributes, neighborhood feature aggregates, and structural encodings; in \graphpfn{}, raw features are processed by graph-attention adapters. For a query node $v$, the prediction is expressed as
\begin{equation}
\hat y_v \;=\; f\!\left(r(v)\;\middle|\;\{(r(u),y_u): u\in \cS\}\right),
\end{equation}
where $\cS\subseteq L$ denotes the labeled \emph{context}. The full-context \gtfm{} and \graphpfn{} protocols use $\cS=L$, that is, they present the complete training table to the backbone. The adapter-based \graphpfn{} protocol additionally materializes the whole graph in accelerator memory so that graph-attention layers can exchange messages across edges.

\subsection{Cost Model and the Memory Wall}
A single forward pass over a context of size $|\cS|$ and $q$ queries produces $\Theta(|\cS|^2 + q\,|\cS|)$ attention-score interactions per layer, or $\Theta((|\cS|^2 + q\,|\cS|)d)$ arithmetic at row width $d$, with activation memory growing correspondingly. Adapter-based backbones further retain all $n$ node tokens, although their actual VRAM also depends on feature width, edge structure, batching, and implementation details. Corresponding resource measurements are presented in Section~\ref{sec:rq2}. The pertinent scaling difference is that whole-graph input residency increases with $n$, while \method{} limits the materialized input through the labeled, query, and halo budgets $k$, $q$, and $h$. On the tabular side, the published G2T-TabPFNv2 protocol employs PCA truncation on two of the eight benchmarks since full-feature preprocessing exhausts memory~\cite{g2tfm}. For the full-attention PFN backbones examined here, expanding model capacity alone does not eliminate this bottleneck: the attention cost continues to be quadratic in the prompt size.

The two budgets regulate separate resources. The labeled budget $k$ governs the amount of supervised evidence visible to the PFN and specifies the length of the labeled context. The halo budget $h$ regulates label-masked structural support for graph adapters and consequently the resident subgraph and its message-passing workload. Raising $h$ does not introduce labeled demonstrations. Raising $k$ does not inevitably restore missing neighbors for a query or context node, since the chosen labeled nodes need not complete its local receptive field. Combining the two into one node budget would consequently obscure their distinct statistical and systems roles.

\subsection{Problem: Budgeted Context Construction}
\begin{definition}[Context construction]
For a query cluster $\cQ_c\subseteq\cQ$, a labeled budget $k$, and a halo budget $h$, a \emph{context policy} $\pi$ chooses (i) a labeled context $\cS(\cQ_c)\subseteq L$ with $|\cS|\le k$, and (ii) a label-masked halo $\cH(\cQ_c)\subseteq\cV\setminus(\cQ_c\cup\cS(\cQ_c))$ with $|\cH|\le h$; the backbone subsequently predicts on the subgraph induced by $\cQ_c\cup\cS\cup\cH$.
\end{definition}
The halo constitutes the graph-specific component of the policy. Each halo label is masked, including when the node belongs to $L\setminus\cS$, such that a halo node provides features and edges but no supervision and consumes no label capacity. Its purpose is to complete, or approximate, the receptive fields of context and query tokens for backbones with layers that attend across edges. The tabular special case ($h{=}0$, no graph) corresponds to retrieval-augmented tabular ICL ~\cite{localpfn,tabdpt}. The graph setting introduces structural retrieval signals, the halo resource, and the cluster-sharing opportunity employed in Section~\ref{sec:method}.

This definition covers the main comparison protocols as special cases. In the row-based setting, $h{=}0$, $k{=}|L|$, and $\cS(\cQ_c){=}L$ reproduce full-context inference. Choosing the query-cluster size as $q{=}1$ yields per-query retrieval, whereas $q{>}1$ yields a cluster-shared context. For an adapter backbone, when $\cQ_c\cup\cS\cup\cH{=}\cV$ and $\cS{=}L$, the materialized input and visible labels correspond to the full-graph protocol. Budgeted context construction lies in the intermediate regime: it constrains labeled evidence and, where required, structural support without altering the frozen backbone.

An effective policy should equal or surpass full-context accuracy at $k\ll|L|$, constrain peak accelerator memory during backbone inference through the query, context, and halo budgets instead of whole-graph residency, amortize retrieval and encoding among queries, operate unchanged with both pure-tabular and adapter-based backbones, and need no training or tuning beyond standard held-out model selection. The following section develops \method{} according to these requirements.

\section{\method{}: Graph-Aware Context Construction}
\label{sec:method}

\method{} builds a bounded prompt for every graph-local group of queries. It initially groups query nodes such that a single retrieved context can support several nearby queries (Section~\ref{sec:clustering}). It subsequently fills the labeled budget through one of three retrieval channels (Section~\ref{sec:channels}) and, for edge-aware backbones, incorporates unlabeled nodes that maintain local message passing (Section~\ref{sec:halo}). The channel and budget are chosen using held-out data without test labels (Section~\ref{sec:auto}).

\subsection{Graph-Local Query Clustering}
\label{sec:clustering}

LoCalPFN creates an individual local kNN context for every query~\cite{localpfn}. On graphs, a direct per-query strategy is costly: every query would need a distinct retrieval step and backbone forward. \method{} instead arranges the query nodes according to a reverse Cuthill--McKee (RCM) permutation of the adjacency matrix. RCM is a bandwidth-reduction heuristic that generally positions graph-adjacent nodes close to each other in the resulting order~\cite{cuthillmckee}. We divide this order into clusters containing $q$ queries; every cluster obtains one retrieved context and one backbone forward. This decreases the number of forwards from $|\cQ|$ to $\lceil|\cQ|/q\rceil$ while maintaining graph locality among the queries in a cluster. The RCM order is calculated once and cached, and any balanced graph partitioner could substitute for the clustering step. Concurrent tabular research independently introduces cluster-batched contexts for efficient inference using distribution-matched retrieval ~\cite{crumb}. In the graph setting, clustering further specifies the structural center of the retrieval scores in Section~\ref{sec:channels} and the subgraph completed through the halo in Section~\ref{sec:halo}.

\subsection{Retrieval Channels}
\label{sec:channels}

For a query cluster $\cQ_c$ and labeled budget $k$, \method{} allocates a score to every labeled node in $L$ and populates the context from the top of a selected ranking. The following three channels represent distinct assumptions regarding the location of useful label information.

\textbf{PPR channel (structural locality).} The structural channel evaluates a labeled node $u$ using its personalized PageRank mass relative to the query cluster,
\begin{equation}
\begin{aligned}
&s_{\mathrm{ppr}}(u) = \pi_{\cQ_c}(u), \\
&\pi_{\cQ_c} = \alpha\, r_{\cQ_c}
  + (1-\alpha)\, P^{\!\top}\pi_{\cQ_c}.
\end{aligned}
\label{eq:ppr}
\end{equation}
where $P$ denotes the row-stochastic transition matrix of $\cG$, $r_{\cQ_c}$ denotes the uniform distribution over $\cQ_c$, and $\alpha$ represents the teleport probability governing how frequently the walk restarts from the query cluster. We obtain the solution to Eq.~\eqref{eq:ppr} through power iteration; the iteration count determines the accuracy of the fixed-point approximation. The resulting score vector relies solely on the graph and the cluster, and is therefore computed once, cached across policies and budgets, and reused whenever the query cluster remains unchanged. This channel captures the inductive bias that labels are locally smooth across edges.

\textbf{Feature-kNN channel (attribute similarity).} The attribute channel evaluates a labeled node according to the negative distance from its closest query row,
\begin{equation}
s_{\mathrm{knn}}(u) \;=\; -\min_{v\in\cQ_c}\,
\bigl\lVert \tilde x_u - \tilde x_v \bigr\rVert_2,
\label{eq:knn}
\end{equation}
calculated over the transformed feature rows $\tilde x$ processed by the backbone. This channel directly transfers retrieval-augmented tabular ICL~\cite{localpfn,tabdpt} to the graph setting, and functions as the feature-space control in our channel ablations.

\textbf{Coverage channel (global representativeness).} The coverage channel employs a seeded uniform sample of $L$ as its basic form (the \texttt{random} row of the results tables). Stratified variants sample proportionally across classes, or across target quantiles for regression. This channel is suitable when global coverage represents the target better than graph locality. It further explains why a budgeted context may occasionally surpass the full table: uniform sampling retains the label marginal in expectation, stratified sampling explicitly preserves class or target-quantile coverage, and both may exclude rows unrelated to the current cluster.

The deployed method and its primary ablations employ these three channels. When a channel cannot satisfy a cluster's budget, for instance because PPR mass is restricted to a disconnected component, the coverage channel supplies the remainder.

\subsection{The Unlabeled Halo}
\label{sec:halo}

For adapter-based backbones (\graphpfn{}), tokens communicate information across edges in the materialized subgraph. A context constructed solely from rows shortens the receptive field of every token whose neighbors were not retrieved. \method{} consequently incorporates nodes from the 1-hop closure of $\cQ_c \cup \cS(\cQ_c)$ as a label-masked halo, capped at $h$ and sampled uniformly beyond the cap. Every node in $\cQ_c\cup\cS\cup\cH$ enters the materialized graph-adapter subgraph, although the PFN interface handles the three sets differently: only $\cS$ is presented as labeled context, only $\cQ_c$ is gathered as queries, and each node in $\cH$ is label-masked. This mask remains applied when a halo node belongs to $L\setminus\cS$, ensuring that membership in the global labeled pool cannot disclose its target following retrieval.

Halo nodes consequently use memory but no label capacity. Their feature rows and induced edges stay resident within the tensors processed by the graph-adapter blocks, enabling query and context tokens to obtain messages from halo neighbors. They are never converted into labeled demonstrations and are omitted from evaluation outputs. The sole path from $\cH$ to a prediction is via feature- and edge-dependent adapter messages, rather than a visible label or supervised loss term. In the halo-budget ablations, performance rises as $h$ grows before reaching saturation. On tolokers-2, incorporating the halo lifts the subgraph protocol from substantially below the full-graph result to parity with it. On artnet-exp, Fig.~\ref{fig:halo} in Section~\ref{sec:rq3} demonstrates that it lifts the subgraph protocol $0.27$ AP beyond the published full-graph accuracy. When the full 1-hop closure remains within $h$, first-layer neighborhoods are retained exactly; otherwise the sampled halo provides a bounded approximation.

One possible explanation is that maintaining local receptive fields in a subgraph whose scale more closely resembles the graphs employed during backbone pretraining yields a more favorable input than whole-graph inference.

\begin{algorithm}[t]
\caption{\method{} inference}
\label{alg:logic}
\small
\KwIn{graph $\cG$, labeled set $L$ with $y_L$, queries $\cQ$, channel $\pi$, budgets $k, h$, cluster size $q$, frozen backbone $f$}
$\sigma \leftarrow$ RCM order of $\cG$;\quad
partition $\cQ$ by $\sigma$ into clusters $\{\cQ_c\}$ of size $q$\;
\ForEach{cluster $\cQ_c$}{
  score $L$ by channel $\pi$ (PPR mass from $\cQ_c$ / min feature distance
  / coverage sampling); $\cS \leftarrow$ top-$k$ (coverage-filled)\;
  $\cH \leftarrow$ label-masked 1-hop closure of $\cQ_c \cup \cS$, capped at $h$
  \tcp*{adapter backbones}
  $\hat y_{\cQ_c} \leftarrow f\big(\text{rows}(\cQ_c) \mid
  \text{rows}(\cS),y_{\cS};\ \cG[\cQ_c \cup \cS \cup \cH]\big)$\;
}
\KwOut{predictions $\hat y_\cQ$}
\end{algorithm}

\subsection{Channel and Budget Configuration}
\label{sec:auto}

The retrieval-channel comparison in Section~\ref{sec:rq3} demonstrates that no individual channel performs best across all datasets because the channels represent distinct assumptions regarding where label information resides. For the reported \method{} configurations, we assess a coarse $\times 4$ budget ladder using held-out validation data and preserve one (channel, $k$) pair for each dataset. The identical runs generate the budget curves; no test labels are employed and no parameters are fitted. When several candidates lie within one standard deviation of the highest held-out score, they constitute a tie set. These ties frequently arise when budgets saturate and channel rankings converge. We resolve ties through measured edge-label assortativity: distinctly positive assortativity favors PPR, whereas distinctly negative assortativity favors coverage. Around zero, or if the favored channel is missing from the tie set, the candidate with the highest score is preserved. Feature-kNN thus remains available for selection whenever it achieves the strongest held-out score, and the tie-break never supersedes a clear score margin. Table~\ref{tab:ablation} verifies that the highest-performing fixed channel differs among the eight benchmarks, although held-out and test rankings may still diverge on particular datasets.

\subsection{Complexity Analysis and Implementation}
\label{sec:cost}

Let $n$ and $m$ denote the numbers of nodes and edges, $L$ the labeled pool, $\cQ$ the query set, $q$ the cluster size (thus $C = \lceil |\cQ|/q \rceil$ clusters), $k$ the label budget, $h$ the halo cap, and $d$ the row width.

Preprocessing consists of one RCM pass over the adjacency ($O(n+m)$) together with, for \gtfm{} featurization, the host system's own encodings. Both are calculated once for each graph and stored on disk. For every cluster, retrieval requires either one batched PPR power iteration ($O(\text{iters}\cdot m)$) or one blocked distance scan across the labeled pool ($O(|L|\,d)$). PPR scores are stored across budgets and reused whenever the cluster remains unchanged. Stratified coverage requires an additional $O(|L|)$ bookkeeping. Over all $C$ clusters, the associated retrieval bounds are $O(C\,\text{iters}\cdot m)$ and $O(C|L|d)$. In implementation, PPR and coverage retrieval execute on the CPU, whereas feature-kNN employs a blocked scan on the accelerator. Retrieved induced subgraphs are constructed from cached sparse adjacency data, preventing reconstruction of the entire graph for every cluster. The context policy stays restricted to the inference path and does not modify backbone parameters.

The protocols vary primarily in inference cost, as summarized in Table~\ref{tab:complexity}. Under the PFN attention pattern employed by both host systems, the $\kappa$ context rows attend inside the context and every one of the $q$ queries reads the context; queries do not attend to each other. A forward consequently incurs $\Theta((\kappa^2+q\kappa)d)$. The full-context protocol uses $\kappa = |L|$ in every one of its $C$ forwards, per-query retrieval uses $|\cQ|$ forwards with $\kappa = k$, and shared contexts need only $C$ forwards with $\kappa = k$. For a row-independent backbone, resident rows decrease from $|L|+q$ to $k+q$. The full-graph adapter protocol further retains all $n$ graph nodes and their edges, while the retrieved-subgraph protocol limits resident nodes to $k+q+h$. Actual VRAM further depends on induced edges, row width, and model activations; Table~\ref{tab:system} measures the resulting decrease relative to whole-graph inference.

\begin{table}[t]
\centering
\caption{PFN attention cost and row residency of full-context, per-query, and cluster-shared protocols.}
\label{tab:complexity}
\scriptsize
\resizebox{\columnwidth}{!}{%
\begin{tabular}{lccc}
\toprule
Protocol & Forwards & Attention / forward & Peak residency \\
\midrule
Full context / graph & $C$ & $\Theta((|L|^2+q|L|)d)$ & $|L|+q$ \\
Per-query retrieval & $|\cQ|$ & $\Theta((k^2+k)d)$ & $k+1$ \\
\method{} (shared) & $C$ & $\Theta((k^2+qk)d)$ & $k+q$ \\
\bottomrule
\end{tabular}
}
\end{table}

The implementation incorporates the context policies into the evaluation pipelines of the host systems while keeping pretrained checkpoints and training procedures unchanged. In the full-context or full-graph setting, it retains the original preprocessing, ensembling, and evaluation metrics from each host system.

\section{Experiments}
\label{sec:exp}

We structure the evaluation around three questions. \textbf{RQ1} contrasts the predictive performance of the deployed \method{} configuration with per-dataset trained models and published graph foundation models on GraphLand. \textbf{RQ2} evaluates efficiency, peak memory, scalability, and cross-backbone transfer relative to backbone-matched full-context or full-graph systems. \textbf{RQ3} separates the retrieval channels, budget, halo, and measurable graph properties that explain the conditions under which each component helps.

\subsection{Experimental Setup}

We conduct evaluations on eight GraphLand benchmarks~\cite{graphland} covering 12k--168k nodes, classification (AP) and regression (R\textsuperscript{2}), and both assortative and disassortative targets. An additional 1.63M-node pokec-regions experiment examines scale beyond the primary benchmark suite. We employ three frozen backbone configurations belonging to two families: pure-tabular \tabpfn{} and LimiX backbones with the \gtfm{} graph featurization, and the adapter-based \graphpfn{}-1.3. The predictive evaluation includes all three configurations, whereas the resource study concentrates on \tabpfn{} and \graphpfn{}. The eight GraphLand benchmarks employ graph-local query clustering, five seeds, and the random-label (RL) splits. TabPFNv2 experiments preserve complete feature rows except in the avazu-ctr comparison, which adopts the PCA-64 preprocessing from the corresponding G2T-FM protocol.

\begin{table}[t]
\centering
\caption{GraphLand datasets and RL-split statistics. $|\cL|$ denotes the labeled context pool; edges count undirected pairs; Assort.\ denotes train-edge label assortativity.}
\label{tab:datasets}
\scriptsize
\setlength{\tabcolsep}{2.6pt}
\resizebox{\columnwidth}{!}{%
\begin{tabular}{lrrrrllc}
\toprule
Dataset & Nodes & Edges & Feat. & $|\cL|$ & Task & Metric & Assort. \\
\midrule
tolokers-2 & 11{,}758 & 519{,}000 & 16 & 1{,}175 & bin.\ cls. & AP & $-0.06$ \\
artnet-exp & 50{,}405 & 280{,}348 & 75 & 5{,}040 & bin.\ cls. & AP & $+0.01$ \\
artnet-views & 50{,}405 & 280{,}348 & 50 & 5{,}040 & regr. & R$^2$ & $+0.23$ \\
city-roads-M & 57{,}073 & 107{,}104 & 26 & 3{,}610 & regr. & R$^2$ & $+0.58$ \\
hm-prices & 46{,}563 & 10{,}730{,}995 & 41 & 4{,}656 & regr. & R$^2$ & $+0.14$ \\
twitch-views & 168{,}114 & 6{,}797{,}557 & 4 & 16{,}811 & regr. & R$^2$ & $-0.39$ \\
city-reviews & 148{,}801 & 1{,}165{,}415 & 37 & 13{,}887 & bin.\ cls. & AP & $+0.11$ \\
avazu-ctr & 76{,}269 & 10{,}984{,}077 & 260 & 7{,}626 & regr. & R$^2$ & $+0.38$ \\
pokec-regions & 1{,}632{,}803 & 22{,}301{,}964 & 11 & 163{,}430 & 183-cls. & Acc & $+0.43$ \\
\bottomrule
\end{tabular}%
}
\end{table}

Budgeted policies reuse preprocessing artifacts via content-keyed caches, such that comparisons within a backbone modify only the context. We present mean$\pm$std across five seeds for the principal predictive and channel comparisons, and employ exact nonparametric tests for seed-level claims. Runtime is evaluated end-to-end for every inference configuration, and memory refers to peak allocated accelerator memory. Resource values are contrasted exclusively within backbone-matched experimental blocks. Prior to evaluating context policies, we validated both host pipelines on our hardware with locally runnable reference configurations. Table~\ref{tab:predictive} contrasts \method{} with published predictive baselines, whereas Table~\ref{tab:system} provides representative performance and resource comparisons between the original protocols and the respective LoGIC configurations.

\subsection{Predictive Performance on GraphLand (RQ1)}
\label{sec:rq1}

\begin{table*}[t]
\centering
\caption{Predictive performance on the GraphLand RL splits (mean$\pm$std). Published results are marked by $\dagger$. Within each backbone-matched pair, bold marks the higher mean. The LoGIC configurations are selected without using test labels and report five seeds.}
\label{tab:predictive}
\scriptsize
\setlength{\tabcolsep}{1.25pt}
\begin{tabular}{llcccccccc}
\toprule
Backbone / model & Inference & \rotatebox{60}{tolokers-2} &
\rotatebox{60}{artnet-exp} & \rotatebox{60}{artnet-views} &
\rotatebox{60}{city-roads-M} & \rotatebox{60}{hm-prices} &
\rotatebox{60}{twitch-views} & \rotatebox{60}{city-reviews} &
\rotatebox{60}{avazu-ctr} \\
& Metric & AP & AP & R$^2$ & R$^2$ & R$^2$ & R$^2$ & AP & R$^2$ \\
\midrule
\multicolumn{10}{l}{\emph{Per-dataset trained references}~\cite{g2tfm}}\\
LightGBM-NFA$\dagger$ & Trained &
56.34{\tiny$\pm$0.06} & 46.13{\tiny$\pm$0.04} &
56.10{\tiny$\pm$0.02} & 61.18{\tiny$\pm$0.03} &
70.84{\tiny$\pm$0.04} & 60.14{\tiny$\pm$0.01} &
78.53{\tiny$\pm$0.01} & 31.71{\tiny$\pm$0.01} \\
GCN$\dagger$ & Trained &
56.27{\tiny$\pm$0.31} & 44.86{\tiny$\pm$0.36} &
56.03{\tiny$\pm$0.25} & 58.82{\tiny$\pm$0.25} &
68.02{\tiny$\pm$0.42} & 75.51{\tiny$\pm$0.05} &
77.81{\tiny$\pm$0.15} & 32.00{\tiny$\pm$0.16} \\
GraphSAGE$\dagger$ & Trained &
54.43{\tiny$\pm$0.34} & 45.14{\tiny$\pm$0.36} &
49.32{\tiny$\pm$0.91} & 59.44{\tiny$\pm$0.27} &
70.00{\tiny$\pm$0.74} & 66.29{\tiny$\pm$0.32} &
78.17{\tiny$\pm$0.10} & 31.44{\tiny$\pm$0.16} \\
GAT$\dagger$ & Trained &
57.41{\tiny$\pm$0.85} & 45.06{\tiny$\pm$0.52} &
53.60{\tiny$\pm$0.24} & 59.86{\tiny$\pm$0.20} &
72.07{\tiny$\pm$1.22} & 72.89{\tiny$\pm$0.27} &
77.74{\tiny$\pm$0.21} & 32.63{\tiny$\pm$0.17} \\
GT$\dagger$ & Trained &
56.98{\tiny$\pm$0.55} & 46.41{\tiny$\pm$0.71} &
53.37{\tiny$\pm$0.46} & 59.55{\tiny$\pm$0.28} &
69.44{\tiny$\pm$0.94} & 72.13{\tiny$\pm$0.13} &
77.34{\tiny$\pm$0.21} & 31.11{\tiny$\pm$0.49} \\
\addlinespace
\multicolumn{10}{l}{\emph{Backbone-matched frozen inference}}\\
\multirow{2}{*}{G2T-TabPFNv2~\cite{g2tfm}}
& Full context$\dagger$ &
\textbf{60.39}{\tiny$\pm$0.19} & 45.73{\tiny$\pm$0.03} &
59.72{\tiny$\pm$0.14} & 60.12{\tiny$\pm$0.03} &
\textbf{65.75}{\tiny$\pm$0.02} & 70.21{\tiny$\pm$0.04} &
77.28{\tiny$\pm$0.25} & \textbf{28.00}{\tiny$\pm$0.36} \\
& LoGIC (ours) &
60.18{\tiny$\pm$0.15} & \textbf{45.94}{\tiny$\pm$0.04} &
\textbf{59.74}{\tiny$\pm$0.06} & \textbf{60.14}{\tiny$\pm$0.12} &
65.37{\tiny$\pm$0.44} & \textbf{73.25}{\tiny$\pm$0.05} &
\textbf{77.54}{\tiny$\pm$0.09} & 27.55{\tiny$\pm$0.01} \\
\addlinespace[1pt]
\multirow{2}{*}{G2T-LimiX~\cite{g2tfm}}
& Full context$\dagger$ &
\textbf{61.60}{\tiny$\pm$0.18} & 48.42{\tiny$\pm$0.78} &
61.58{\tiny$\pm$0.08} & 65.16{\tiny$\pm$0.07} &
76.14{\tiny$\pm$0.08} & 71.31{\tiny$\pm$0.06} &
\textbf{78.98}{\tiny$\pm$0.44} & \textbf{32.70}{\tiny$\pm$0.14} \\
& LoGIC (ours) &
61.45{\tiny$\pm$0.22} & \textbf{48.87}{\tiny$\pm$0.14} &
\textbf{61.78}{\tiny$\pm$0.07} & \textbf{65.17}{\tiny$\pm$0.06} &
\textbf{76.20}{\tiny$\pm$0.14} & \textbf{74.15}{\tiny$\pm$0.01} &
76.97{\tiny$\pm$0.57} & \textbf{32.70}{\tiny$\pm$0.18} \\
\addlinespace[1pt]
\multirow{2}{*}{GraphPFN~\cite{graphpfn}}
& Full graph$\dagger$ &
61.29{\tiny$\pm$0.12} & 51.79{\tiny$\pm$0.11} &
\textbf{62.79}{\tiny$\pm$0.08} & 64.85{\tiny$\pm$0.13} &
77.88{\tiny$\pm$0.09} & \textbf{73.20}{\tiny$\pm$0.08} &
\textbf{80.25}{\tiny$\pm$0.05} & 31.63{\tiny$\pm$0.06} \\
& LoGIC (ours) &
\textbf{61.35}{\tiny$\pm$0.05} & \textbf{52.06}{\tiny$\pm$0.18} &
62.65{\tiny$\pm$0.08} & \textbf{64.88}{\tiny$\pm$0.15} &
\textbf{77.89}{\tiny$\pm$0.13} & 72.12{\tiny$\pm$0.05} &
80.22{\tiny$\pm$0.05} & \textbf{32.17}{\tiny$\pm$0.31} \\
\bottomrule
\end{tabular}
\end{table*}

Table~\ref{tab:predictive} shows the primary predictive comparison on GraphLand. The first block lists individually trained LightGBM-NFA and GNN references from the G2T-FM evaluation. The second block positions every full-context or full-graph result alongside its backbone-matched LoGIC configuration, separating the effect of context construction from a backbone change. For LoGIC-TabPFNv2, the channel and budget are selected among the structural, feature, and coverage candidates according to Section~\ref{sec:auto}. The candidate set contains a terminal budget that depletes the labeled pool when that pool is smaller than the following budget step. The LimiX and GraphPFN rows likewise employ LoGIC configurations chosen without test labels. For LimiX, the candidates comprise the published 10-member ensemble and a terminal budget that depletes the labeled pool; the chosen terminal-budget configurations are applied to tolokers-2, artnet-views, city-roads-M, hm-prices, and avazu-ctr. These rows examine whether the context-construction interface transfers among backbones instead of conducting a per-backbone oracle search. Fixed channels are not shown as independent proposed methods here; RQ3 employs them as controlled ablations.

With the same \tabpfn{} backbone, \method{} preserves full-context accuracy while using a bounded context. Wherever our full-context reproduction can run, $k{=}4096$ equals it within run-to-run variation or surpasses it. On artnet-exp, \method{} exceeds the published G2T-TabPFNv2 result, increasing from $45.73$ to $45.94$. The labeled pools for tolokers-2, artnet-views, and city-roads-M are depleted at the terminal candidate. For the larger pools, LoGIC surpasses the published G2T-TabPFNv2 result on twitch-views and city-reviews. On avazu-ctr, the protocol-matched PCA-64 configuration obtains $27.55{\pm}0.01$ R$^2$, with a $0.45$-point gap relative to the published ten-member mean.

In aggregate, LoGIC-TabPFNv2 numerically matches or surpasses G2T-TabPFNv2 on five of the eight datasets. The residual mean gaps equal $0.21$ AP on tolokers-2, $0.38$ R$^2$ on hm-prices, and $0.45$ R$^2$ on avazu-ctr.

The transferred LimiX configurations surpass their published full-context counterpart on five of eight datasets and match it on avazu-ctr. In addition to raising artnet-exp from $48.42$ to $48.87$ AP and twitch-views from $71.31$ to $74.15$ R$^2$, the chosen configurations attain $61.78{\pm}0.07$ R$^2$ on artnet-views, $65.17{\pm}0.06$ R$^2$ on city-roads-M, and $76.20{\pm}0.14$ R$^2$ on hm-prices. The reported avazu-ctr means coincide at two-decimal precision ($32.70$ R$^2$). The GraphPFN transfer surpasses the published full-graph result on artnet-exp by $0.27$ AP and marginally surpasses it on tolokers-2, city-roads-M, hm-prices, and avazu-ctr. It remains within $0.14$ R$^2$ on artnet-views and $0.03$ AP on city-reviews, whereas twitch-views retains a $1.08$ R$^2$ gap. Collectively, these rows substantiate the conclusion that LoGIC generally maintains the predictive performance of the respective full-context or full-graph protocol across backbone families, while surpassing it on multiple datasets.

The external rows bound the claim. \method{}-TabPFNv2 surpasses the strongest individually trained baseline on tolokers-2 and artnet-views, falls within $0.47$ AP on artnet-exp, and lags behind the strongest trained result by $1.04$ R$^2$ on city-roads-M and $0.99$ AP on city-reviews, with wider gaps on hm-prices, twitch-views, and avazu-ctr. Similarly, full-graph \graphpfn{} and G2T-LimiX lead on multiple datasets when large-memory inference is accessible. The contribution consequently does not constitute a universal accuracy substitute for per-dataset training. Instead, it is a budgeted frozen-inference protocol that retains competitive performance while eliminating the full-context resource constraint, as evaluated next.

\subsection{Efficiency, Scalability, and Cross-Backbone Comparison (RQ2)}
\label{sec:rq2}

\begin{table*}[t]
\centering
\caption{Performance, inference time, and peak allocated VRAM of the full-context/full-graph baselines and deployed LoGIC configurations (mean$\pm$std).}
\label{tab:system}
\scriptsize
\setlength{\tabcolsep}{4.5pt}
\begin{tabular}{lllrrr}
\toprule
Backbone & Dataset & Method & Performance & Time & Peak VRAM \\
\midrule
\multirow{4}{*}{TabPFNv2}
& tolokers-2 & G2T-TabPFNv2 (full context) & $60.04{\pm}0.07$ AP & $6.84{\pm}0.70$\,s & 4.10\,GB \\
&              & LoGIC (ours) & $60.18{\pm}0.15$ AP & $7.66{\pm}0.89$\,s & 2.20\,GB \\
& artnet-exp   & G2T-TabPFNv2 (full context) & $45.68{\pm}0.03$ AP & $117.78{\pm}1.37$\,s & 6.90\,GB \\
&              & LoGIC (ours) & $45.94{\pm}0.04$ AP & $65.68{\pm}0.80$\,s & 5.80\,GB \\
\midrule
\multirow{4}{*}{GraphPFN}
& artnet-exp & GraphPFN (full graph) & $51.86{\pm}0.10$ AP & $39.65{\pm}0.16$\,s & 19.59\,GB \\
&             & LoGIC (ours) & $52.06{\pm}0.18$ AP & $30.49{\pm}0.49$\,s & 8.24\,GB \\
& city-roads-M & GraphPFN (full graph) & $64.86{\pm}0.14$ R$^2$ & $14.69{\pm}0.01$\,s & 7.71\,GB \\
&              & LoGIC (ours) & $64.88{\pm}0.15$ R$^2$ & $11.02{\pm}0.05$\,s & 5.73\,GB \\
\bottomrule
\end{tabular}
\end{table*}

Table~\ref{tab:system} presents local reproductions and resource measurements obtained on our hardware. With \tabpfn{}, LoGIC lowers artnet-exp end-to-end time from $117.78{\pm}1.37$ to $65.68{\pm}0.80$ seconds per seed and peak VRAM from $6.9$ to $5.8$\,GB, while marginally increasing AP.

\begin{figure}[t]
\centering
\includegraphics[width=0.92\linewidth]{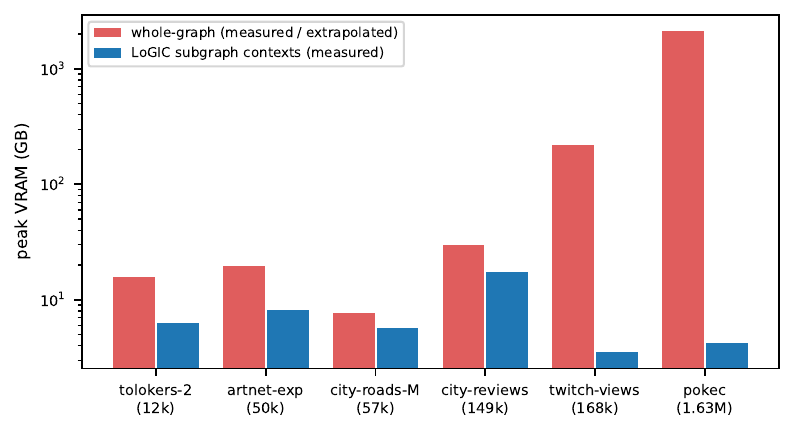}
\caption{Peak VRAM of whole-graph \graphpfn{} inference versus LoGIC subgraph contexts. Whole-graph values use local measurements where available (tolokers-2, artnet-exp, city-roads-M, and city-reviews) and per-node extrapolation for the remaining scale cases; LoGIC values are measured.}
\label{fig:memwall}
\end{figure}

The transfer to \graphpfn{} evaluates another bottleneck: full-graph residency for adapter message passing. LoGIC processes one retrieved induced subgraph for each query cluster, without altering or retraining the checkpoint. In the matched local comparison on artnet-exp, it achieves $52.06{\pm}0.18$ AP using 8.24\,GB, versus $51.86{\pm}0.10$ AP and 19.59\,GB under full-graph inference. On city-roads-M, LoGIC attains $64.88{\pm}0.15$ R$^2$ using 5.73\,GB, versus $64.86{\pm}0.14$ R$^2$ and 7.71\,GB under full-graph inference. Figure~\ref{fig:memwall} illustrates how this separation expands with graph size.

\paragraph{Million-node scale study}
Pokec-regions comprises 1.63M nodes, 44.6M directed edges, and 183 classes. LoGIC predicts 76.8k cluster-sampled held-out queries with $36.3\%$ accuracy in 40 minutes using 4.2\,GB peak VRAM. Whole-graph residency would demand approximately two terabytes at the observed rate. A random context with the same $k{=}4096$ achieves only $8.6\%$, suggesting that retrieval continues to be important even when the labeled budget constitutes merely $0.25\%$ of the graph. This experiment serves as a scale demonstration instead of a same-split SOTA claim. Posterior calibration is evaluated independently below.

\subsection{Ablation and Mechanistic Analysis (RQ3)}
\label{sec:rq3}

\paragraph{Retrieval channels}
\begin{table*}[t]
\centering
\caption{Retrieval-channel ablation on the GraphLand RL splits using frozen TabPFNv2. All channels use full-feature preprocessing and $k{=}4096$. Bold marks the best fixed-channel result per dataset.}
\label{tab:ablation}
\scriptsize
\setlength{\tabcolsep}{3.5pt}
\begin{tabular}{lcccccccc}
\toprule
Variant & \rotatebox{60}{tolokers-2} & \rotatebox{60}{artnet-exp} &
\rotatebox{60}{artnet-views} & \rotatebox{60}{city-roads-M} &
\rotatebox{60}{hm-prices} & \rotatebox{60}{twitch-views} &
\rotatebox{60}{city-reviews} & \rotatebox{60}{avazu-ctr} \\
\midrule
Coverage/random only & \textbf{60.22} & 45.44 & 59.33 & 60.11 &
65.37 & \textbf{72.00} & 74.25 & 23.42 \\
Feature channel only & 60.20 & 44.81 & 59.36 & 60.11 &
\textbf{65.43} & 68.95 & 71.67 & \textbf{24.37} \\
Structural (PPR) only & 60.20 & \textbf{45.94} & \textbf{59.48} &
\textbf{60.14} & 65.32 & 70.20 & \textbf{75.28} & 23.96 \\
\bottomrule
\end{tabular}
\end{table*}

Table~\ref{tab:ablation} positions the three candidate channels in their designated roles. PPR represents the graph-structural channel, feature-kNN constitutes the retrieval-based tabular ICL control, and coverage/random serves as the structure-agnostic control. No fixed channel performs best across all datasets; the strongest channel varies among datasets. This finding motivates dataset-dependent channel selection instead of a single universal retriever.

Table~\ref{tab:assortativity} connects fixed-channel outcomes with train-edge label assortativity. Positive assortativity frequently coincides with a strong PPR result, while the negatively assortative twitch-views graph prefers coverage/random sampling. The relationship is empirical rather than deterministic: near-zero assortativity cannot account for the PPR advantage on artnet-exp, and strong node attributes render feature-kNN best on avazu-ctr.

\begin{table}[t]
\centering
\caption{Train-edge label assortativity and the best fixed retrieval channel at $k{=}4096$. Regression uses Spearman correlation; classification uses the same-class edge rate minus chance.}
\label{tab:assortativity}
\small
\setlength{\tabcolsep}{4pt}
\begin{tabular}{lcc}
\toprule
Dataset & Assort. & Best fixed channel \\
\midrule
tolokers-2 & $-0.057$ & coverage/random \\
artnet-exp & $+0.007$ & PPR \\
artnet-views & $+0.228$ & PPR \\
city-roads-M & $+0.583$ & PPR \\
hm-prices & $+0.139$ & feature-kNN \\
twitch-views & $-0.388$ & coverage/random \\
city-reviews & $+0.113$ & PPR \\
avazu-ctr & $+0.378$ & feature-kNN \\
\bottomrule
\end{tabular}
\end{table}

\begin{figure*}[t]
\centering
\includegraphics[width=\textwidth]{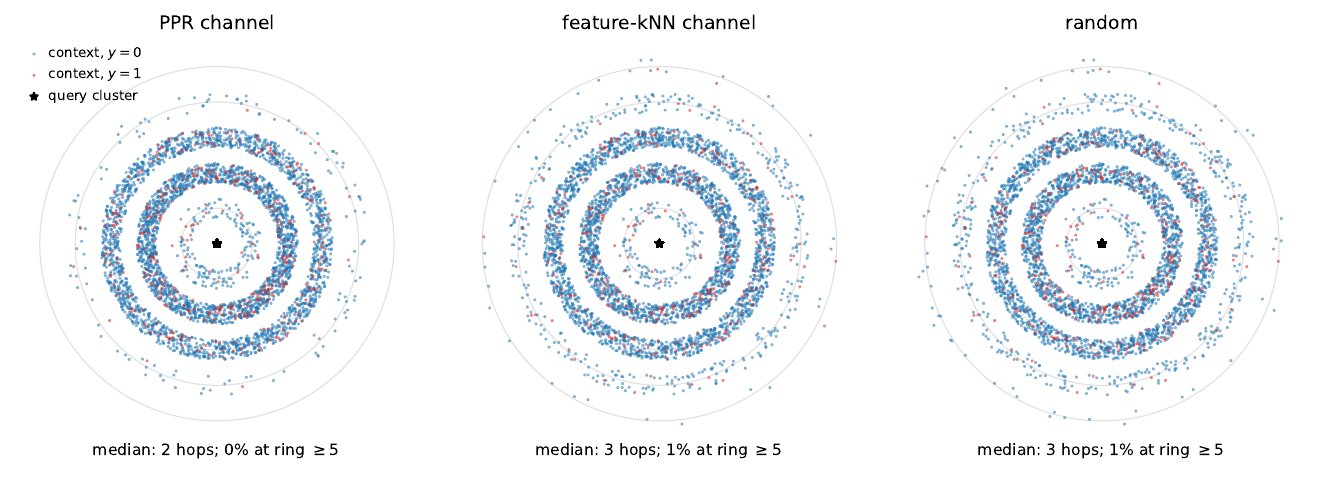}
\caption{Retrieved-context geometry for one deterministically selected artnet-exp query cluster at $k{=}4096$. Radius is BFS hop distance from the query cluster; angle is a deterministic node hash used only to separate markers. Color denotes the context label and the central marker denotes the query cluster.}
\label{fig:anatomy}
\end{figure*}

\paragraph{Retrieved-context geometry}
Figure~\ref{fig:anatomy} displays what the three fixed channels retrieve for a single deterministically selected artnet-exp query cluster. Radius denotes the BFS hop distance from the query cluster; angle denotes a deterministic node hash employed solely to distinguish markers. The PPR context exhibits greater graph locality, with a median distance of two hops and no retrieved nodes at radius five or above. Feature-kNN and random sampling both yield a median distance of three hops and position 1\% of their contexts at radius five or above. This evidence verifies the structural-locality bias of PPR, although it does not indicate that PPR will constitute the most accurate channel on every graph.

\begin{figure*}[t]
\centering
\includegraphics[width=\textwidth]{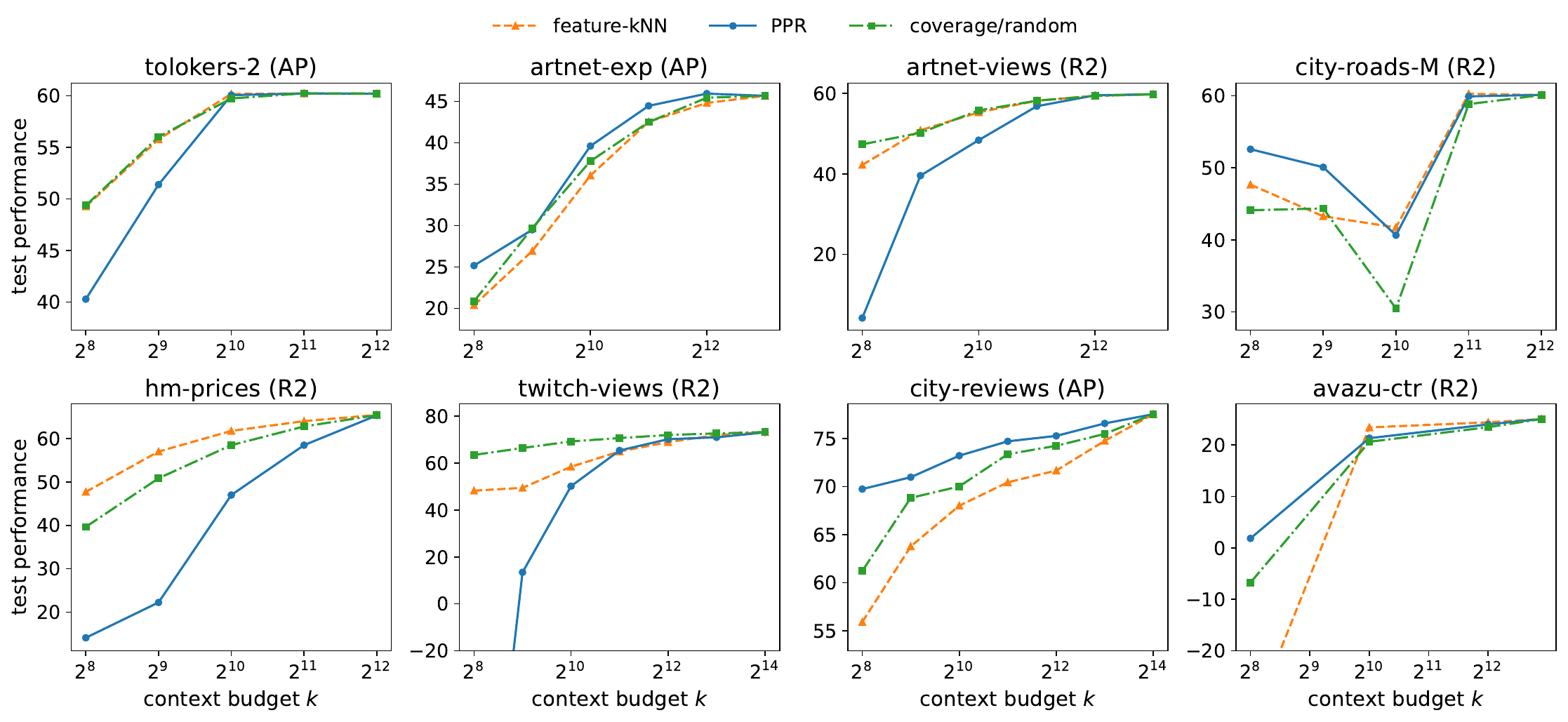}
\caption{Test performance versus labeled-context budget $k$ for the three LoGIC channels. The first seven panels show budgets available for all three channels. In the avazu-ctr panel, the final displayed budget $|\cL|{=}7{,}626$ is the complete labeled pool; the three channels coincide when the pool is exhausted.}
\label{fig:budget}
\end{figure*}

\paragraph{Budget regimes}
Figure~\ref{fig:budget} demonstrates that every channel deteriorates below approximately 512 rows, while larger budgets reveal dataset-dependent channel orderings and ultimately approach saturation. This behavior motivates choosing both the retrieval channel and the context budget. Table~\ref{tab:sensitivity} independently alters the PPR teleport probability and query ordering. artnet-exp remains stable over the tested $\alpha$ values, whereas city-reviews shows greater sensitivity to locality; in the reported comparison, RCM ordering further outperforms natural node order. Consequently, the qualitative channel conclusions are not contingent on a single PPR setting, although retrieval configuration may still be important for individual datasets.

\begin{table}[t]
\centering
\caption{Sensitivity of the PPR channel ($k{=}4096$) to teleport $\alpha$ under RCM query ordering, with natural node order as a reference. Values report test AP (mean$\pm$std over five seeds).}
\label{tab:sensitivity}
\tiny
\setlength{\tabcolsep}{1.5pt}
\resizebox{\columnwidth}{!}{%
\begin{tabular}{lccccc}
\toprule
 & $\alpha{=}0.15$ (def.) & $\alpha{=}0.05$ & $\alpha{=}0.30$ & $\alpha{=}0.50$ & nat.\ order \\
\midrule
artnet-exp & 45.94\tiny$\pm$0.04 & 46.00\tiny$\pm$0.06 & 45.93\tiny$\pm$0.05 & 45.91\tiny$\pm$0.06 & 45.93\tiny$\pm$0.03 \\
city-reviews & 75.28\tiny$\pm$0.05 & 74.95\tiny$\pm$0.04 & 75.77\tiny$\pm$0.01 & 75.90\tiny$\pm$0.04 & 74.38\tiny$\pm$0.02 \\
\bottomrule
\end{tabular}
}
\end{table}

\begin{figure}[t]
\centering
\includegraphics[width=0.9\linewidth]{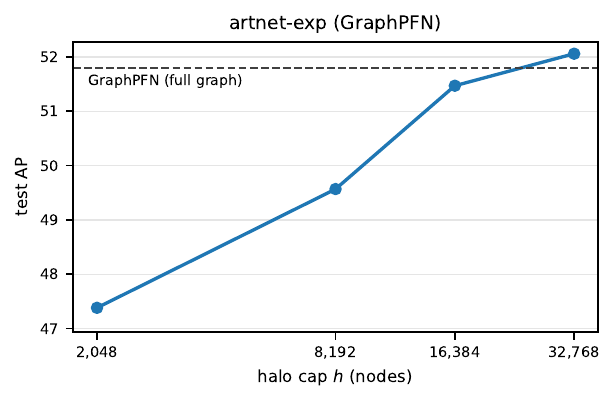}
\caption{Halo ablation on artnet-exp (\graphpfn{}): test AP versus the halo cap, with the published GraphPFN full-graph score as a reference.}
\label{fig:halo}
\end{figure}

\paragraph{Unlabeled halo}
For adapter backbones, a context derived solely from labeled rows shortens the graph-attention neighborhood. When the 1-hop closure of $\cQ_c\cup\cS$ remains within the halo cap, incorporating its label-masked nodes retains the first-layer neighborhoods of query and context tokens; a finite cap provides a sampled approximation. Figure~\ref{fig:halo} quantifies this effect. On artnet-exp, enlarging the cap increases AP to $52.06{\pm}0.18$, surpassing the published full-graph result by $0.27$ AP. On tolokers-2, the respective fixed-PPR experiment at $k{=}4096$ increases from $54.20$ AP without a halo to $61.09$ AP after the 1-hop support is incorporated, a gain of $6.89$ AP that matches the full-graph reference.

\paragraph{Posterior calibration}
Table~\ref{tab:calibration-compact} contrasts representative full and budgeted contexts through Brier score and adaptive expected calibration error (ECE), where smaller values signify better probabilistic predictions. Brier score quantifies squared probability error, whereas adaptive ECE contrasts confidence with empirical accuracy across 15 equal-mass bins. On tolokers-2, substituting feature-kNN for the full context shifts Brier score from $12.17$ to $12.15$ and ECE from $1.60$ to $1.59$. On artnet-exp, the PPR context increases AP from $45.68$ to $45.94$, while Brier score declines from $7.06$ to $7.03$ and ECE from $1.51$ to $0.85$. Therefore, in these representative settings, the memory reduction achieved through budgeting yields no apparent calibration penalty. This audit makes no claim of consistent calibration gains across channels or datasets.

\begin{table}[!htbp]
\centering
\caption{Representative posterior calibration results using frozen G2T-TabPFNv2 (mean$\pm$std; adaptive ECE with 15 equal-mass bins). Lower Brier and ECE are better.}
\label{tab:calibration-compact}
\scriptsize
\setlength{\tabcolsep}{2.2pt}
\begin{tabular}{llccc}
\toprule
Dataset & Context & AP & Brier$\times100$ & ECE$\times100$ \\
\midrule
tolokers-2 & Full context & $60.04{\pm}0.07$ & $12.17{\pm}0.02$ & $1.60{\pm}0.01$ \\
 & Feature-kNN & $60.10{\pm}0.10$ & $12.15{\pm}0.01$ & $1.59{\pm}0.20$ \\
artnet-exp & Full context & $45.68{\pm}0.03$ & $7.06{\pm}0.00$ & $1.51{\pm}0.03$ \\
 & PPR & $45.94{\pm}0.04$ & $7.03{\pm}0.01$ & $0.85{\pm}0.06$ \\
\bottomrule
\end{tabular}
\end{table}

\section{Discussion}
\label{sec:discussion}

\subsection{Deployment Considerations}

Our findings indicate several deployment considerations for PFN-class graph foundation models.

The initial choice concerns the labeled budget. When the local full-context protocol is runnable, budgeted contexts with $k{=}4096$ maintain its performance using bounded memory and comparable or lower latency. For larger labeled pools, they stay competitive with published full-context results. The complete table consequently does not need to serve as the default prompt. Whether extra rows enhance or diminish performance depends on the dataset. \mbox{Figure~\ref{fig:budget}} demonstrates this behavior: on artnet-exp, the bounded context surpasses the locally reproduced full-context result by $0.26$ AP. Budgets beneath the capacity floor ($k_0\!\approx\!512$ rows for current backbones) should not be used because every evaluated selection policy performs weakly in this regime.

The second decision involves the retrieval channel. Train-edge label assortativity can be calculated in a single sparse pass and helps guide this decision. When assortativity is distinctly positive, PPR retrieval provides a strong default, with feature-kNN as another candidate when raw attributes are informative. When assortativity is distinctly negative, coverage sampling is typically preferable, since local retrieval may be detrimental at small budgets. When candidates lie within one standard deviation of the highest held-out score, the selection rule in Section~\ref{sec:auto} employs measured assortativity to resolve the tie without superseding a clear score margin.

The remaining resource decision relates to structural support. For adapter backbones, enlarging the label budget does not inevitably enhance performance, whereas assigning memory to the unlabeled halo recovers local receptive fields. Figure~\ref{fig:halo} demonstrates this effect: the resulting subgraph inference matches full-graph inference on tolokers-2, while LoGIC surpasses the published GraphPFN full-graph result by $0.27$ AP on artnet-exp.

\subsection{Scope and Future Work}
\label{sec:limitations}

The evaluation covers transductive node classification and regression on graphs with tabular node attributes, which is the setting in which the host systems are defined and evaluated. Retrieval uses the graph neighborhood around each query, and the GraphLand benchmarks make that neighborhood available at inference time. Inductive and streaming graphs, whose neighborhoods arrive progressively, together with link- and graph-level tasks, extend the same two-resource formulation and are natural next steps.

The reported configuration keeps one retrieval channel and one context budget per dataset through held-out model selection, so no test labels enter the procedure. The three channels are deliberately simple and training-free, and budgeted contexts bound the input presented to the backbone rather than the total workload, since runtime still grows with the number of query clusters. For graph-adapter backbones, the halo restores the 1-hop neighborhoods of query and context tokens within its cap. Learned mixtures of retrieval signals, cluster-adaptive budgets, transfer rules that reuse configurations among graphs, and multi-hop halos tied to adapter depth could each strengthen this layer while preserving the inference-only interface of \method{}.

More generally, larger backbones may accommodate larger contexts and render the observed capacity floor less restrictive, but an admissible prompt must still contain a specific selection of rows. We consequently anticipate that context construction will remain pertinent to this model family, together with feature construction and backbone design, and will extend naturally to inductive and dynamically evolving graphs.

\section{Conclusion}
\label{sec:conclusion}

This work identifies context construction as a key inference-time design problem for node-level graph ICL with frozen tabular foundation models. We express the problem as allocation across two bounded resources: labeled evidence for the PFN and label-masked structural support for graph adapters. Based on this formulation, \method{} integrates graph-local query sharing with complementary structural, feature-based, and coverage retrieval channels, and chooses a channel--budget configuration without test labels. The resulting interface keeps the host backbone and its training procedure unchanged while constraining PFN attention and peak input residency, eliminating the need to store the entire labeled pool or the whole graph in accelerator memory.

Across GraphLand benchmarks and three backbone configurations drawn from two model families, bounded contexts preserve competitive predictive performance while lowering the computational and memory burden associated with full-context and full-graph inference, including at million-node scale. The experiments further demonstrate that retrieval depends on the graph: structural locality, attribute similarity, and global coverage are beneficial under distinct regimes. Context construction should consequently adjust to measurable graph properties rather than depend on one retrieval rule across all datasets. These findings establish context construction as a general modeling and systems layer for graph foundation models. Future research should examine its interaction with fine-tuning, learned mixtures of retrieval signals, inductive and streaming graphs, link- and graph-level ICL, and halo policies matched to adapter receptive fields.

\IEEEtriggeratref{28}
\bibliographystyle{IEEEtran}
\bibliography{references}

\end{document}